\documentclass[10pt, a4paper]{article}
\usepackage{lrec2022} 
\usepackage{multibib}
\newcites{languageresource}{Language Resources}
\usepackage{graphicx}
\usepackage{tabularx}
\usepackage{soul}
\usepackage{todonotes}
\usepackage{float}
\usepackage{enumitem}
\usepackage[greek,english]{babel}
\usepackage{titlesec}

\usepackage{epstopdf}
\usepackage[utf8]{inputenc}

\usepackage{hyperref}
\usepackage{xstring}

\usepackage{color}

\usepackage{tabularx}

\usepackage{booktabs}
\usepackage{multirow}
\usepackage{makecell}
\usepackage{newtxtext,newtxmath}
\usepackage{makecell,tabularx}
\usepackage{siunitx}

\titleformat{\section}{\normalfont\large\bfseries\center}{\thesection.}{1em}{}
\titleformat{\subsection}{\normalfont\SmallTitleFont\bfseries\raggedright}{\thesubsection.}{1em}{}
\titleformat{\subsubsection}{\normalfont\normalsize\bfseries\raggedright}{\thesubsubsection.}{1em}{}
\renewcommand\thesection{\arabic{section}}
\renewcommand\thesubsection{\thesection.\arabic{subsection}}
\renewcommand\thesubsubsection{\thesubsection.\arabic{subsubsection}}

 \renewcommand\theadfont

\title{Cross-Lingual Knowledge Transfer for Clinical Phenotyping }

\name{Jens-Michalis Papaioannou$^{\ast}$, Paul Grundmann$^{\ast}$, Betty van Aken$^{\ast}$, Athanasios Samaras$^{\dagger}$  \\ {\bf \large Ilias Kyparissidis$^{\ddagger}$, George Giannakoulas$^{\dagger}$, Felix Gers$^{\ast}$, {\bf \large Alexander L\"oser$^{\ast}$}}}

\address{
         $^{\ast}$DATEXIS, Berliner Hochschule für Technik (BHT), Germany,\\
         $^{\dagger}$First Department of Cardiology, AHEPA University Hospital, Aristotle University of Thessaloniki, Greece\\ 
         $^{\ddagger}$Laboratory of Medical Physics and Digital Innovation, Aristotle University of Thessaloniki, Greece, \\
         \{michalis.papaioannou, paul.grundmann, bvanaken,  gers, aloeser\}@bht-berlin.de\\
         \{asamaraa, ggiannakoulas\}@auth.gr\\
        \{iliaskypkok\}@for.auth.gr
        }

\abstract{
Clinical phenotyping enables the automatic extraction of clinical conditions from patient records, which can be beneficial to doctors and clinics worldwide. However, current state-of-the-art models are mostly applicable to clinical notes written in English. We therefore investigate cross-lingual knowledge transfer strategies to execute this task for clinics that do not use the English language and have a small amount of in-domain data available. We evaluate these strategies for a Greek and a Spanish clinic leveraging clinical notes from different clinical domains such as cardiology, oncology and the ICU. Our results reveal two strategies that outperform the state-of-the-art: Translation-based methods in combination with domain-specific encoders and cross-lingual encoders plus adapters. We find that these strategies perform especially well for classifying rare phenotypes and we advise on which method to prefer in which situation. Our results show that using multilingual data overall improves clinical phenotyping models and can compensate for data sparseness. 
 \\ \newline 
 }
\begin{document}

\maketitleabstract

\section{Introduction}

\paragraph{Clinical phenotyping from text.}
Clinical information extraction has the potential to support clinicians in their daily work. Phenotyping—the extraction of patient conditions from text—in particular, can help clinicians to summarize patient states and to find similar patients more easily. Recent years have seen many advancements in automatic phenotyping from clinical text \cite{pheno_bert_mulyar,zhang-etal-2021-self}, most of them based on neural networks. Especially domain-specific pre-training of such networks has shown to be of great benefit in clinical information extraction \hbox{\cite{Li2020BEHRTTF,pubmedbert}}. However, current state-of-the-art models are mostly exclusively applicable to English text, due to the large amount of both labeled and unlabeled clinical text resources in English. 

\paragraph{The need for cross-lingual transfer.}  We want to investigate how to leverage the clinical knowledge encoded in English and other languages to the benefit of (low-resource) target languages. This scenario is commonly described as cross-lingual knowledge transfer \cite{Ponti2020XCOPAAM}. Medicine is increasingly transformed by globalisation \cite{Labont2011TheGI} and knowledge collected in different languages has significant impact on local decisions (as seen during the Covid-19 pandemic). However, cross-lingual knowledge transfer in the clinical domain remains largely unexplored. Our work takes the perspective of clinicians working with non-English clinical notes. 
In particular we evaluate on medical texts based on Spanish and Greek in which there are far less such resources openly available.


\paragraph{Which cross-lingual approach works best in the \hbox{clinical} domain?}
Differences between languages (and clinics) regarding patient distributions, documentation styles or typologies make clinical knowledge transfer challenging. 
We evaluate three current approaches for cross-lingual transfer to find out which method handles these challenges best: 1) Translating Greek and Spanish notes into English before using a medical-specific English encoder, comparable to work by \newcite{use_machine_translation}. 2) Using multi-lingual encoders \cite{XLMR}, which allow knowledge transfer between languages, but are not pre-trained on medical data. 3) Expanding multi-lingual encoders with adapters \cite{madX}, which have shown promising results on other cross-lingual tasks \cite{Le2021LightweightAT} while being more resource efficient.

\paragraph{Cross-lingual transfer for rare phenotypes.}
Some phenotypes are more common in some parts of the world than in others. Distributing such knowledge to other regions is an important goal of medical knowledge transfer. Cross-lingual transfer is especially promising for these scenarios in which few examples exist in the target language. Therefore, we  analyse in particular how the long-tail of phenotypes benefits from all evaluated cross-lingual approaches.

\paragraph{Contributions.}
\begin{enumerate}
    \item{We compare state-of-the-art methods for cross-lingual knowledge transfer regarding their performance in the clinical domain. To this end, we analyse whether the incorporation of clinical notes written in different languages improves the results of clinical phenotyping.}
    \item{We evaluate methods on three datasets in Spanish, Greek and English (for transfer only) and infer which methods are best suited for different scenarios when working with non-English clinical notes. Our analysis further reveals which combinations of languages are most beneficial.}
    
    \item{We show in particular that the performance on the long tail of phenotypes benefits from the usage of additional data in different languages.}
\end{enumerate}
We publish the code for all experiments to ensure \hbox{reproducibility}\footnote{\url{https://github.com/neuron1682/cross-lingual-phenotype-prediction}}.

\begin{figure*}[t]
   \includegraphics[width=2\columnwidth]{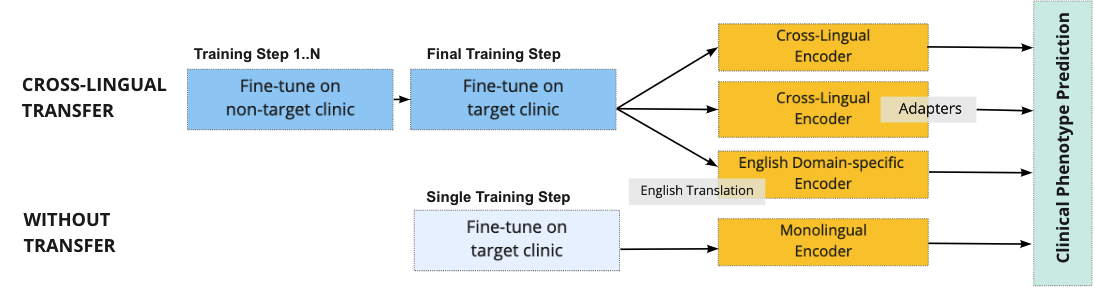}
  \caption{Schematic demonstration of our approaches. We compare cross-lingual with monolingual approaches. For the knowledge transfer we use sequential transfer learning starting from Mimic (high-resource dataset). We distinguish between cross-lingual encoders, cross-lingual encoders plus adapters and English domain-specific encoders with prior translation.}
\label{fig:methodology}
\end{figure*}

\section{Related Work}

\paragraph{Phenotyping from clinical notes.}
Clinical phenotyping is the task of extracting patient conditions from Electronic Health Record (EHR) data. In this work, we focus on clinical notes--discharge summaries in particular--as information source, since they  summarize the current state of a patient. \newcite{pheno_rules_solt} proposed
rule-based algorithms to solve the task, while \newcite{pheno_cnn_yao,pheno_cnn_wang} applied Convolutional Neural Networks. Recently, large pre-trained language models have shown to outperform these earlier approaches \cite{pheno_bert_mulyar}. However, such language models are mostly used to process English clinical text data, for which large corpora such as PubMed Central exist. In contrast, we examine ways of solving the task in languages without a variety of openly available medical text data and evaluate how to transfer (clinical) knowledge from one language to another.

\paragraph{Models with cross-lingual abilities.}
In recent years, a variety of multi- and cross-lingual language models have been introduced \cite{Xtreme_benchmark}. For this work, cross-lingual models are of special interest due to their inherent ability to transfer knowledge between languages \cite{yarowski}. To this end, \newcite{chung-etal-2020-improving} propose to cluster and merge the vocabularies of similar languages for joint learning. The LASER model \cite{laser} is trained on parallel data of 93 languages with a shared BPE vocabulary. XLM \cite{XLM} additionally pre-trains BERT with parallel data. Building on these approaches, the XLM-R model \cite{XLMR}, pre-trained on 100 languages, performs especially well on low-resource languages, which leads us to examine whether it is also of use in the clinical domain. \newcite{madX} introduced language adapters to cross-lingual models. Using adapters has shown to improve performance on multiple tasks while being more parameter-efficient. Thus, we investigate whether they are also beneficial to cross-lingual clinical phenotyping.

\paragraph{Language transfer via translations.}
In contrast to model-based approaches, \newcite{use_machine_translation} propose the use of translations for knowledge transfer between languages. While \newcite{translationArtifacts} highlight that translations can have a regulating effect, they also show how translation errors can accumulate throughout the pipeline. These contrasts lead us to evaluate both the use of cross-lingual and translation based approaches to improve clinical phenotyping of low-ressource datasets.

\paragraph{Cross-lingual knowledge transfer in the medical domain.}
While there is a number of monolingual pre-trained language models for the medical domain in languages other than English \cite{spanish_clinicalbert}, the effect of cross-linguality on medical text has been rarely studied. 
\newcite{schafer_multilingual_icd} enrich a translated version of the Spanish CodiEsp dataset with English data to improve the performance in ICD code prediction. \newcite{brasilian_biobert} use a pre-trained multi-lingual BERT model to further train on Brasilian clinical notes. We are the first to conduct an extensive analysis over multiple cross-lingual approaches regarding their usefulness for the clinical domain.

\section{Task and Datasets}

This section describes the clinical phenotyping task and introduces the three datasets used for evaluating cross-lingual transfer methods.

%

\subsection{Phenotyping Task}
The goal of clinical phenotype prediction is to support medical doctors to automatically categorise patients by the clinical conditions reported in a clinical note. This task is crucial e.g.,~for understanding the full picture of occurring diseases at a clinic, enabling further studies using specific patients, and predicting disease development and outcome. 
Here, we use the pre-defined CCSR\footnote{\url{https://hcup-us.ahrq.gov/toolssoftware/ccsr/ccs_refined.jsp}} categorization where each category represents a set of ICD codes. Each category may represent a disease or a set of e.g.,~different arrhythmias or ill defined diseases. 
\subsection{Datasets}
We use three datasets namely MIMIC-III \citelanguageresource{mimiciii}, CodiEsp \citelanguageresource{codiesp} and AHEPA-cardio. These datasets consist of clinical notes that originate from different clinics in different countries i.e.~USA, Spain and Greece. The languages in which the notes are written belong to the same language family (indo-european) but are classified into different language branches (Germanic, Italic and Greek). 
Greek is also typologically different from the other two. 

Furthermore, we choose medical data from diverse medical contexts. MIMIC-III originates from the ICU, CodiEsp from different clinics and AHEPAcardio from the cardiology department. The narrative and style of the texts is therefore different. This is also reflected in the length of the texts (see Table \ref{table:note_stats}).
Our rationale is that the medical knowledge in these datasets is diverse and complementary.
 
 
\paragraph{Mimic III - English Language.}
We use the freely-available MIMIC-III v1.4 database (referred to as \textit{Mimic} in the following). It contains de-identified Electronic Health Records (EHR) data including clinical notes in English from the Intensive Care Unit (ICU) of Beth Israel Deaconess Medical Center in Massachusetts between 2001 and 2012. We focus our work on discharge summaries in particular and the diagnosis information associated with an admission. Similar to previous work \cite{clinical_outcome} we use discharge summaries and only keep the most informative sections regarding the phenotyping task including \textit{Chief Complaint}, \textit{History of Present Illness}, \textit{Physical Exam}, \textit{Social/Family History} and \textit{Brief Hospital Course}. We further filter out re-admissions, duplicates and notes about newborns.

\paragraph{CodiEsp - Spanish Language.}
The CodiEsp dataset consists of 1,000 clinical case studies manually selected by doctors and cover a diverse set of medical specialties, including oncology, urology, cardiology, pneumonology and infectious diseases. The case studies are annotated regarding the diagnoses mentioned in the text. 
The notes are provided in both the original Spanish language and an English translation. For the translation of the notes the authors used a machine translation system which was fine-tuned on biomedical data \cite{biomedical_translator_codie}.

\paragraph{AHEPAcardio - Greek Language.}
The AHEPAcardio dataset (referred to as \textit{Ahepa} in the following) is a collection of around 2,400 discharge summaries and originates from the cardiology clinic of the AHEPA University Hospital in Greece.\footnote{The publication of parts of the data is currently under review.} The discharge summaries date from 2013 to 2020 and are annotated regarding clinical diagnoses found in the text by a team of medical doctors. 

\paragraph{Mappings to 334 CCSR Target Labels.}
We map  all three datasets into a common label space of CCSR codes using the official mapping\footnote{\url{https://hcup-us.ahrq.gov/toolssoftware/ccsr/DXCCSR_v2021-2.zip}}.  We limit our CCSR target space to labels in the Spanish dataset CodiEsp, since it has the biggest variety of CCSR codes. This results in 334 clinical disease categories for CodiEsp, from which 60 appear in Ahepa and 326 in Mimic. We use iterative stratification \cite{multilabel_stratified} to split each of the datasets in train/dev/test depending on the medical conditions of the patients. This leads to a representative group of patients in each of the splits for every clinic separately. In Table \ref{table:note_stats} we note the number of clinical notes i.e., the number of patients in each of the sets. 
Note that not all labels appear in each of the splits. The occurence of labels ranges from one to thousand.
For example, several types of cancers appear only once and medical conditions that appear most are \textit{Other general signs and symptoms}, \textit{Essential Hypertension} and \textit{Respiratory signs and symptoms} in CodiEsp, Mimic and Ahepa respectively. 


\begin{table}[h]
\centering

\begin{tabular}{lrrrc} 
\toprule
\multicolumn{5}{c}{\textbf{Clinical Note Statistics}} \\ 
\midrule
 & \multicolumn{1}{c}{\textbf{Train}} & \multicolumn{1}{c}{\textbf{Dev}} & \multicolumn{1}{c}{\textbf{Test}} & \textbf{Ø Length} \\ 
\midrule
CodiEsp & 656 & 165 & 175 & 351 \\
Ahepa & 1,592 & 402 & 393 & 257 \\
Mimic & 24,758 & 6,187 & 6,182 & 649 \\
\bottomrule
\end{tabular}
\caption{Split details of each dataset: CodiEsp, Ahepa and MIMIC. Average length of each note is measured in words.}
\label{table:note_stats}

\end{table}

%
%

\section{Methods}


The task of clinical phenotyping has been rarely investigated in a multilingual scenario. We compare and discuss approaches for cross-lingual transfer that have already been proven successful in other domains. We restrict our approaches to sequential transfer learning, since it allows to share models across clinics without having to share patient data explicitly. An alternative approach would be to combine different datasets. However, in this case clinics are forced to share their data which is often difficult due to data regulations. 
Fig \ref{fig:methodology} overviews the evaluated approaches.

\paragraph{Translation into English for monolingual models.} The first approach we evaluate is to translate all notes into English and predict phenotypes with monolingual language models. While the translation may result in erroneous results, translation services are widely available. Therefore this approach is a decent baseline for common practice.

First, for the task of translating notes from Greek into English, we use pre-trained Opus-MT translation models\footnote{We are restricted to use open source solutions due to privacy regulations of patient data. The same restrictions apply to clinics.} \cite{opus_mt} via EasyNMT\footnote{\url{https://github.com/UKPLab/EasyNMT}}. The machine translation of CodiEsp is provided with the corpus and was performed with a system adapted to the biomedical domain. Next, we apply PubMedBERT \cite{pubmedbert} a state-of-the-art model trained from scratch on a large medical corpus in English and optimized for tasks like ICD code prediction. As a result, we expect the model to produce higher quality representations for domain specific terms. We fine-tune PubMedBERT sequentially on Mimic and on translated data in different combinations. 





\paragraph{Pre-trained cross-lingual models.}
The second approach is the use of pre-trained cross-lingual models or multi-lingual models with cross-lingual capabilities such as XLM-R \cite{XLMR}. Cross-lingual models do not require to translate the low- resource data into English. They can represent the same information from different languages close to each other in the embedding space. 
In contrast, multilingual models are typically pre-trained on only a language modeling task with data from multiple languages. 
This pre-training schema does not enforce the model to map semantically similar texts close to each other in the embedding space. Nevertheless, \newcite{XLMR} have shown that even multilingual pre-trained models can have cross-lingual properties and even outperform cross-lingual models in cross-lingual downstream tasks.
We use a pre-trained multilingual model with cross-lingual capabilities and fine-tune it sequentially on different clinical datasets. By keeping the label space identical, we force the model to reuse knowledge from one dataset in one language for the next dataset. We first fine-tune on the high-resource dataset and then continue training on the low-resource data. This reflects a scenario of closed data silos \cite{data_silos}, where only the pre-trained model is accessible, but not the training data.

\paragraph{Language and task adapters.}
As a third approach, we use adapters
\cite{madX} to enhance a multilingual model with additional language- and task specific parameters. 
Adapters consist of feed-forward layers that are added to each encoder layer in a transformer model. Typically, only the parameters of the adapters are trained while the rest of the parameters of the base model stay frozen.
In consequence, they are an efficient method to fine-tune models since it involves significantly fewer parameters compared to fine-tuning the entire model.
Furthermore, adapters can prevent catastrophic forgetting, because the model parameters are frozen and only the adapters are trained. Hence it preserves the ability of a cross-lingual model to map similar inputs from different languages close to each other in the same embedding space even though we train with an unequal number of samples in different languages. 
We use pre-trained language adapters from AdapterHub \cite{pfeiffer2020AdapterHub} in combination with a cross-lingual pre-trained model to enhance the representation on the low-resource datasets. By reusing the fine-tuned weights of the task adapter for each dataset, we enforce that knowledge from previous fine-tunings is kept and reused for further trainings. 
In contrast to the task adapters, we keep the language adapters frozen during the training and choose a pre-trained language adapter that matches the language of the training dataset.


\paragraph{Monolingual baseline without knowledge transfer.}
Finally, we use mono-lingual, domain-specific pre-trained language models which provide a baseline for the training without any knowledge transfer. 
Domain-specific language models are usually monolingual due to the sparse data available in the subject domain. They are also more common in high-resource languages. 
In return, the models usually provide an improved representation for technical terms in the specific domain. The most common approach is to train the downstream task on the clinical notes in their native language.
For each dataset, we separately fine-tune a different, state-of-the-art pre-trained language model and measure the performance on each dataset individually.
However, in this scenario, the model only learns the underlying distribution of labels of the given dataset. We therefore expect this approach to perform worse, especially for long-tail or zero-shot predictions.
Furthermore, we also perform experiments with adapters and translation on each target, low-resource dataset in combination with cross-lingual (XLM-R) as well as monolingual pre-trained models as a baseline.

\begin{table*}[ht!]
\centering
\begin{tabularx}{\textwidth}{ Xcc XX} 
\toprule
\multicolumn{1}{l}{\multirow{2}{*}{\textbf{Model}}} & \multicolumn{2}{c}{\textbf{Clinical Phenotyping}}          \\
\multicolumn{1}{c}{}                                & \textbf{Macro-AUC [\%] } & \textbf{Macro PR-AUC [\%] }  \\ 
\midrule
\textbf{Single Dataset Training} \\
Monolingual Spanish BERT (C)                        & 82.00                    & 25.91                        \\
Spanish Biomedical Clinical RoBERTa (C)             & 84.58                    & 29.89                        \\
XLM-R (C)                                           & 56.64                    & 5.28                         \\ 
XLM-R + Adapters (C)                                & 61.96                    & 6.43                         \\
Translation + PubMedBERT (C$_T$)                    & 83.45                    & 29.54                        \\ 
\midrule
\textbf{Multi Dataset Training} \\
XLM-R (M $\rightarrow$ C)                                         & 83.52                    & 25.96                        \\
XLM-R (M $\rightarrow$ A $\rightarrow$ C)                                       & 83.82                    & 25.96                        \\
\hline
XLM-R + Adapters (M $\rightarrow$ C)                              & 85.63                    & 34.41                        \\
XLM-R + Adapters (M $\rightarrow$ A $\rightarrow$ C)                            & 83.90                    & 32.22                        \\ 
\hline
Translation + PubMedBERT (M $\rightarrow$ C$_T$)                  & \textbf{90.95}           & \textbf{43.13}               \\
Translation + PubMedBERT (M $\rightarrow$ A$_T$ $\rightarrow$ C$_T$)            & 90.40                    & 41.98                        \\
\bottomrule
\end{tabularx}
\caption{Performance for \textbf{CodiEsp}. \textit{M}: Mimic, \textit{A}: Ahepa and \textit{C}: CodiEsp. The order represents the fine-tune order. The subscript $_T$ means that the English translation of the texts is used and otherwise the original language. The approach which yields the strongest results is the sequential fine-tuning of the \textbf{Domain specific Encoder} first with Mimic and then with the \textbf{English translation} of CodiEsp. }
\label{table:Codie_results}
\end{table*}

\begin{table*}[ht!]
\centering
\begin{tabularx}{\textwidth}{ Xcc XX} 
\toprule
\multicolumn{1}{l}{\multirow{2}{*}{\textbf{Model}}} & \multicolumn{2}{c}{\textbf{Clinical Phenotyping}}          \\
\multicolumn{1}{c}{}                                & \textbf{Macro-AUC [\%] } & \textbf{Macro PR-AUC [\%] }  \\ 
\midrule
\textbf{Single Dataset Training} \\
Monolingual Greek BERT (A) &  90.18 & 56.22\\
XLM-R (A) &  60.45& 12.31 \\ 
XLM-R + Adapters (A)  &  56.60 &10.30\\
Translation + PubMedBERT (A$_T$) & 83.15 & 37.10\\
\midrule
\textbf{Multi Dataset Training} \\
XLM-R (M $\rightarrow$ A) & 89.87 &50.23\\
XLM-R (M $\rightarrow$ C $\rightarrow$ A)  &  90.03& 51.15\\
\midrule
XLM-R + Adapters (M $\rightarrow$ A) & 90.15 & 54.45\\
XLM-R + Adapters (M $\rightarrow$ C $\rightarrow$ A)  & \textbf{91.50}& \textbf{57.63}\\
\midrule
Translation + PubMedBERT (M $\rightarrow$ A$_T$)  & 86.20& 45.14\\
Translation + PubMedBERT (M $\rightarrow$ C$_T$ $\rightarrow$ A$_T$) &  88.75& 49.90\\
\bottomrule
\end{tabularx}
\caption{Performance for \textbf{Ahepa}. {\textit{M}: Mimic, \textit{A}: Ahepa and \textit{C}: CodiEsp.} {The order represents the fine-tune order.} The subscript $_T$ means that the English translation of the texts is used and otherwise the original language. The approach which yields the strongest results is the sequential fine-tuning of the \textbf{Cross-lingual  Encoder plus Adapter} on Mimic, CodiEsp and Ahepa in \textbf{original language}.}
\label{table:Ahepa_results}
\end{table*}
\subsection{Experimental Setup}
For the monolingual and translation setting, we use PubMedBERT \cite{pubmedbert}. For Spanish, we use Spanish-BERT \cite{SpanishBert} and Spanish Biomedical Clinical RoBERTa \cite{spanish_clinicalbert}. Since no domain specific biomedical or clinical language model for Greek has been released to our knowledge, we use Greek-BERT \cite{GreekBert}, which has been pre-trained on non clinical or biomedical language. 
We use pre-trained models from Huggingface \cite{huggingface}. 
Due to the limited context length of the pre-trained models we truncate all the notes to 512 tokens. 
We perform a Bayesian hyperparameter tuning for learning rate, accumulation gradients, warm-up steps, hidden dropout rate and attention dropout rate. For optimization we use AdamW \cite{AdamW} and for the learning rate we use a learning rate scheduler with warm-up.
We run $50$ trials for each model with $100$ being the maximum number of epochs and use Early Stopping with patience of $5$ epochs. We use AUROC as validation metric for the training.
For each dataset, we perform a hyperparameter optimization and reuse the weights of the best model for the next dataset. This applies for all experiments where sequential fine-tuning is part of the training. In this manner we transfer the knowledge from one dataset to the other. We start training from a high-resource dataset and finally fine-tune on the target dataset on which we evaluate the results. For example, in our results the term Mimic $\rightarrow$ CodieEsp (M $\rightarrow$ C) means that the model was first fine-tuned on Mimic and the next model was initialised with the best performing model on the same dataset and finally fine-tuned on CodiEsp. 


\section{Results}





Tables \ref{table:Codie_results} and \ref{table:Ahepa_results} report the performance on the phenotype classification task on the low-resource datasets CodiEsp and Ahepa, respectively. Each table is separated in two segments, single and multi dataset training. 
In the first segment we report the scores of 1) the monolingual BERT of the respective target language, 2) the cross-lingual model (XLM-R), 3) the cross-lingual model with adapters (XLM-R with adapters) and 4) the English translation in combination with a domain-specific encoder (PubMedBERT). We report the performance of the mentioned models trained with only the target dataset for each task. Additionally, we report for the CodiEsp task the performance of a Spanish domain-specific encoder (Spanish Biomedical Clinical RoBERTa). In the second segment we report the scores of the models we use for cross-lingual knowledge transfer.
We measure the approaches in macro-average AUROC and PR-AUC.



\paragraph{Knowledge transfer surpasses monolingual models.}
For both tasks, knowledge transfer incorporating other non-target datasets yields improved results compared to the monolingual counterparts trained on the target dataset only. Table \ref{table:Codie_results} shows the results for CodiEsp. Interestingly, all knowledge transfer methods outperform all single dataset approaches. Translation in combination with a domain specific encoder works best on the CodiEsp dataset. It outperforms the monolingual Spanish BERT by a large margin (11\%) in AUROC and the uplift for PR-AUC is even higher (62\%). The approach also outperforms the domain-specific Spanish Biomedical Clinical RoBERTa  (by 8\% AUROC and 44\% PR-AUC). 

The cross-lingual encoder in combination with adapters is the best performing approach on Ahepa.
Table \ref{table:Ahepa_results} shows that the best model surpasses the performance of the monolingual Greek BERT. In contrast to the CodiEsp task the monolingual model shows high performance and the use of knowledge transfer results to a more moderate increase in both AUROC ($\approx$2\%) and PR-AUC ($\approx$3\%).

When trained on a single dataset, monolingual models outperform the cross-lingual encoders for both datasets, CodiEsp and Ahepa. This shows that the performance increase is indeed a result of training with clinical datasets in other languages.

\paragraph{Detection of rare conditions improves when using further languages.}

\begin{figure}[t!]
   \includegraphics[width=\columnwidth]{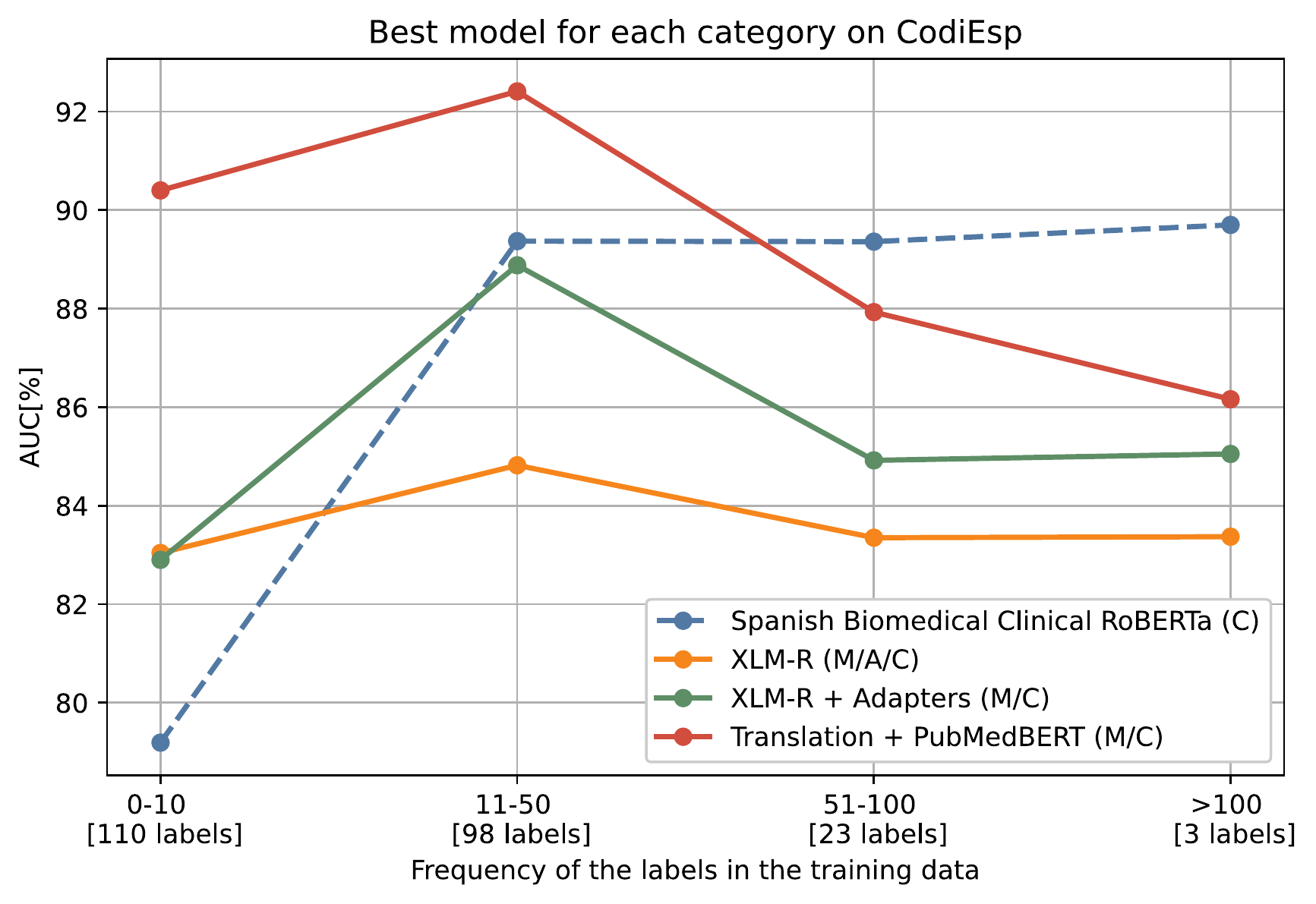}
  \caption{\textbf{CodiEsp}: Performance for different label frequencies in the training data. Baselines using only a single dataset are represented by dashed lines. The number of labels in brackets corresponds to the count of labels of each group in the target test dataset (CodiEsp).}
\label{fig:Nsamples_codie}
\end{figure}

\begin{figure}[t!]
   \includegraphics[width=\columnwidth]{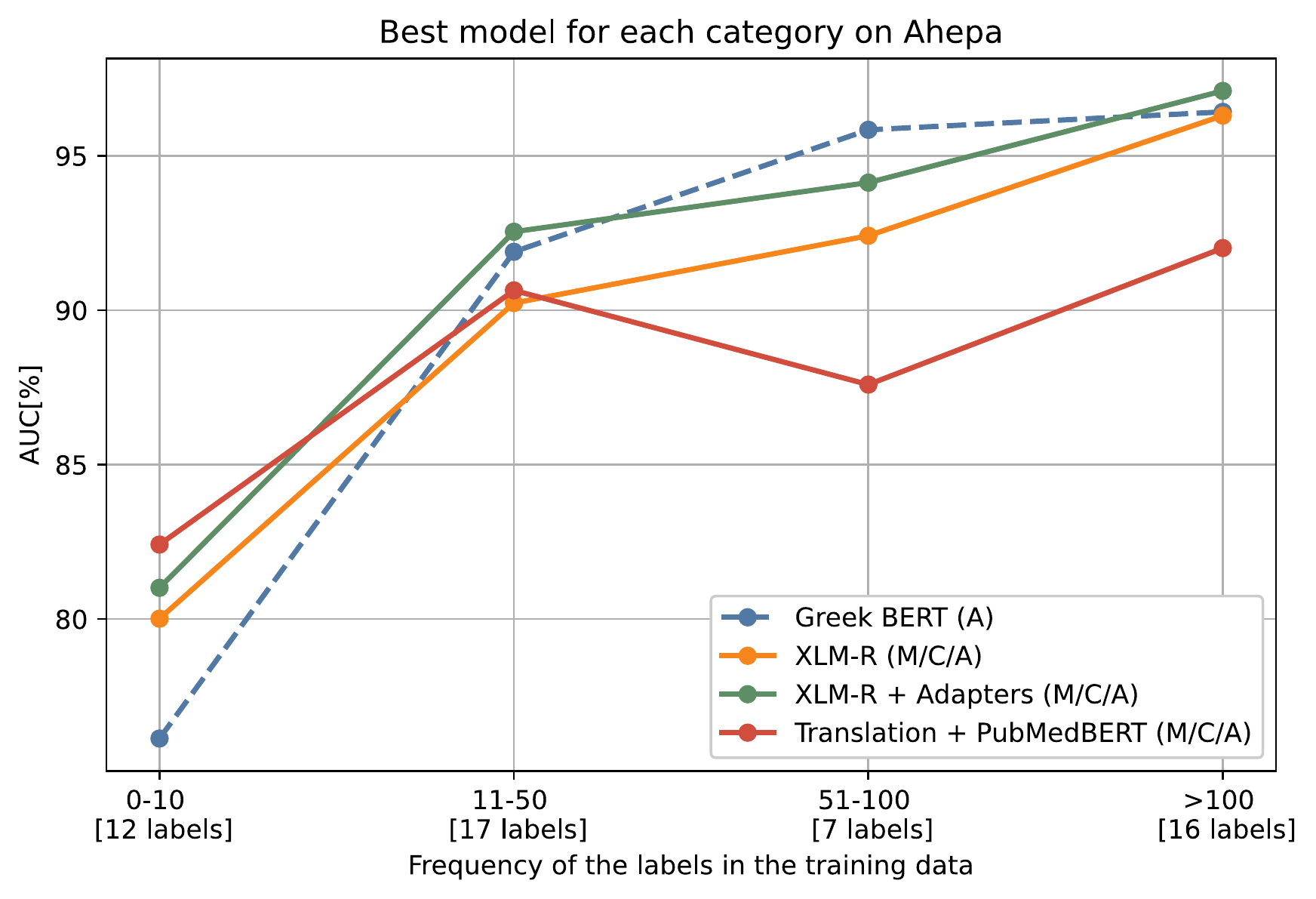}
  \caption{\textbf{Ahepa}: Performance for different label frequencies in the training data. Baselines using only a single dataset are represented by dashed lines. The number of labels in brackets corresponds to the count of labels of each group in the target test dataset (Ahepa).}
\label{fig:Nsamples_achepa}
\end{figure}

To test the performance for rare conditions, we aggregate labels according to their frequency in the training sets and report the macro-averaged AUROC in Figure \ref{fig:Nsamples_codie} and \ref{fig:Nsamples_achepa}. 

For CodiEsp (Fig. \ref{fig:Nsamples_codie}) we observe that approaches using two or three languages work significantly better for low frequency labels (0-50 times) than single language approaches. The increase in AUROC is 13\% in average for the lowest frequency interval (0-10 times). In the interval with medium label frequency (51-100 times) the performance gap decreases. And very frequent labels (>100) are best extracted by a single-language approach.

For Ahepa (Fig. \ref{fig:Nsamples_achepa}) multi-language approaches as well outperform the others regarding very low-frequency labels (0-10). Regarding more frequent labels (11-50, 51-100, >100) the monoligual Greek BERT performs on-par with the cross-lingual Encoder + Adapters approach trained on multiple languages. The differences in AUROC are more moderate than for CodiEsp, but still show that multi-lingual approaches are beneficial for long-tail phenotype detection.

\paragraph{Performance on high-frequency labels.}
In Figure \ref{fig:Nsamples_codie} we observe a drop in performance of the cross-lingual models for the high-frequency labels. The amount of medical categories that fall into the high-frequency groups is very small and they consist of mostly general categories like e.g., \textit{Other specified inflammatory condition of skin} and \textit{Skin/Subcutaneous signs and symptoms}. In contrast the medical conditions in Ahepa are well distributed over the groups (see: number of labels in Fig. \ref{fig:Nsamples_achepa}). In total, 68\% of the codes in the CodiEsp test set benefit from knowledge transfer and 62\% of the Ahepa test set codes.



\paragraph{Adapters are on par with XLM-R full fine-tuning.}
The sequential training of the task adapter on all datasets results are on par or have even better performance compared to the full fine-tuning of the XLM-R model for both datasets. Specifically for CodiEsp, both models perform approximately on par, and in the case of Ahepa it is $\approx$2\% better.


\begin{table*}[htb!]
 \centering
\begin{tabularx}{\textwidth}{XXX}
\toprule
 \textbf{Original text} & \textbf{Failed machine translation} & \textbf{Doctor's translation}\\
 \midrule
\textgreek{Ο ασθενής προσήλθε λόγω\newline \textit{δύσπνοιας} από ημερών} & The patient came due to shortness of days & The patient was admitted due to current \textit{dyspnoea} since days\\
\midrule
\textgreek{ανοικτός βοτάλλειος πόρος} & open shell resource &
 patent ductus arteriosus (PDA) \\
 \midrule
  \textgreek{ΑΥ} & Authorize & Arterial Hypertension \\
  \midrule
  \midrule
  la paciente presenta dolor y \textit{defensa a la palpación profunda} de fosa lumbar derecha  & the patient presented with pain and \textit{defense of the deep fixation} of the right lumbar fossa & patient presented with pain and \textit{defense to deep palpation} of the right lumbar fossa\\
  \midrule 
  realizó estudio \textit{oftalmológico} e inició \textit{con deflazacort} & performed the \textit{diagnostic test}, and initiated treatment \textit{with dementia} & performed an \textit{ophthalmologic} study and started treatment \textit{with deflazacort} \\ 
  \midrule
 no fumadora \textit{ni bebedora de alcohol} &
  non-smoker or \textit{bicuspid bicuspid drinker} & non-smoker and \textit{non-drinker of alcohol} \\
  \bottomrule
  \end{tabularx}
\caption{Text excerpts where machine translation failed. We show examples of both the translation of the Greek and Spanish clinical notes. Note, that the Spanish machine translation system was fine-tuned on biomedical data and the Greek system was not.}
\label{table:failing_translation}
\end{table*}

\section{Discussion}
Our results show that cross-lingual transfer improves performance on the clinical phenotyping task--in particular for low-resource datasets. We further conduct a qualitative error analysis with medical doctors revealing specifics of cross-lingual transfer in the clinical domain. This section discusses selected  results for applying our findings in a clinical setting.  
 \subsection{Strengths and Limitations of Cross-lingual Transfer}

    \paragraph{Knowledge transfer is possible despite dataset \hbox{differences}.} 
        Clinical notes differ from clinic to clinic in a number of ways. They are written in different languages i.e. English, Spanish or Greek. Second, the form of documentation is different. The Mimic and CodiEsp clinical notes are written in a narrative style in contrast to the Ahepa notes which are much denser. The Ahepa notes are in average shorter, use a lot of abbreviations and some parts are not written in full sentences. Additionally, Greek is typologically different to English and Spanish. In spite of those differences, our experiments show that knowledge transfer between those datasets yields better results than the respective monolingual models. We hypothesise that some differences might, in fact, have a positive impact by adding variance to the data.
        
    
    \paragraph{Limitations of cross-lingual knowledge transfer.}
     Tables \ref{fig:Nsamples_codie} and \ref{fig:Nsamples_achepa} show that cross-lingual transfer does not improve performance on all diagnoses groups.
     Our error analysis reveals that the effect of transferring knowledge for the same medical category can have opposite results for the two studied datasets.
     We observe this especially in broad categories such as \textit{Other specified and unspecified skin disorders}. While these work well in Ahepa, the transfer fails for CodiEsp, which might be due to the diverse set of clinics included in CodiEsp resulting in a larger number of diagnoses per category.
     
     We also observe cases in which the monolingual Spanish and Greek encoder almost perfectly recognises relevant clinical cases and knowledge transfer has a negative impact.
     This is the case e.g. for \textit{Pericarditis and pericardial disease}.
      Since every clinic has a different focus, the importance of  clinical facts related to the medical condition is different. This has a direct impact on the amount of details a physician reports about a clinical fact. 
      Consequently, to identify the same medical category different clinical textual cues are needed depending on the clinic it originates from.

    \paragraph{The benefit for rare phenotype detection.} 
    
    Knowledge transfer for rare phenotypes is not only useful for languages with little labeled data but also for countries with smaller populations like Greece. For instance, \textit{Takotsubo cardiomyopathy} is a disease with low occurrence \cite{takotsubo}. Although the frequency of the medical condition may not be different from clinic to clinic, low-resource datasets benefit from datasets with bigger volume of labeled data.
    In our results we observe an increase of 2pp for the category \textit{Other and ill-defined heart disease} to which the condition belongs. Similarly, for the Ahepa task we observe an increase of 2pp for \textit{Acute and unspecified renal failure}, where the cardiology clinic might profit from CodiEsp data from nephrology or pathology clinics. In both cases we observe that the data augmentation improves results for both low-resource datasets because of the higher text volume and added variance.

\subsection{Design Choices in Clinical NLP Setting}
        \paragraph{Adapters and translation are suitable methods for cross-lingual knowledge transfer.}
        Our results suggest that both techniques, adapters and translation to English, are suitable ways for clinical knowledge transfer. 
        In the case of CodiEsp the translation to English and sequentially fine-tuning a large  English domain-specific encoder yields the best results.
        For the Ahepa case, the pre-trained cross-lingual model with adapters trained with clinical notes in native language achieves the best results. In both cases, a transfer method outperforms all other corresponding methods for that data set.
        
        \paragraph{'Rules-of-thumb' considerations.}
        In order to improve the performance of clinical phenotyping some considerations are necessary to pick the best cross-lingual solution. Is there a well performing in-domain translation system\footnote{The system should be deployable within the premises of a clinic due to regulations and sensitivity of patient data.} to English available for the datasets at hand? 
        If this is the case, sequentially fine-tuning a large English domain-specific encoder is a  promising solution.  However, if no high quality machine translation is available,  fine-tuning the adapters in native language is the better choice. In addition, further experiments revealed that the results vary depending on the order with which we fine-tune. Starting the training with the high-resource language mostly yields good results.
        
        \paragraph{Adapters are  efficient for cross-lingual transfer.}  
        The adapter approach proved to be an efficient way for cross-lingual knowledge transfer for the phenotyping task. This comes from the fact that,
        we only train the task adapter parameters.
        So, while the system adds additional parameters to the model to better represent low-resource languages, only a fraction of all the parameters is  updated during training. Similar to \newcite{madX}, we observe on par or better results by training only the task adapter in comparison to the full cross-lingual model in both cases, Ahepa and CodiEsp,  during sequential fine-tuning of those encoders. Thus, in case computational complexity is a limiting factor, a cross-lingual encoder like XLM-R plus adapters is the recommended approach.



\subsection{Additional Observations}
        \paragraph{Mixed results on adding  various languages.}
Tables \ref{table:Codie_results} and \ref{table:Ahepa_results}
compare knowledge transfer methods fine-tuned on two or on three datasets. We found that adding more data does not necessarily improve results. The Ahepa task yields the strongest results when incorporating all three datasets. However, in the case of CodiEsp, we observe that fine-tuning on two languages performs on par or better than on three languages. The best result for the CodiEsp task was achieved using only English and Spanish text instead of additional Greek data. We leave further \hbox{analyses} on which language combinations work best to future work.


\paragraph{Language typology impacts transfer capabilities.} When adding Greek as a third language in the CodiEsp task, knowledge transfer with XLM-R does not improve the performance. It even decreases  when using the adapters approach. This may occur due to the different typologies of the languages and the limited shared vocabulary. Interestingly, when adding Spanish as a third language in the Ahepa task the performance for the adapters approach improves. This may be because the Greek texts contain some standard medical expressions, such as \textit{STEMI}, which are also used in Spanish and English notes.
 
        
\paragraph{Impact of translation quality and consistency.} 
The approach based on English translations yields lower performance increase for Ahepa than for CodiEsp. This might happen because we use two different machine translation systems: A domain-specific translator for CodiEsp and a general one for Ahepa. Translating the same medical terms differently produces different representations and therefore may have a negative or at best no effect on the performance of the downstream task. In Table \ref{table:failing_translation} we report examples of failed translations. Most of the problems we observe with the Greek machine translator are the medical terms and abbreviations doctors frequently use. In the Ahepa (Greek) notes we also notice that sometimes punctuation and accents on words are missing, which can be decisive when translating.

\section{Conclusion}

Our results demonstrate that cross-lingual knowledge transfer improves performance in the clinical phenotype prediction task. In particular we show that knowledge transfer yields the strongest uplift in performance on rare phenotypes. Thus, medical tasks whose targets can be mapped to a single scheme can benefit from adding data in different languages. This is an important result in the medical domain where worldwide standards for describing patient conditions exist. We also give recommendations on how to effectively and efficiently solve such a task that involves notes written in rather distant languages. Our future work will focus on 1) methods specifically targeting clinical cross-lingual transfer and 2) studying the effects of integrating more languages and different machine translation systems. 3) Finally we want to investigate the benefit of cross-lingual transfer for extreme low-resource languages.
%


\section*{Acknowledgments}
We would like to thank Janine Schleicher and Julius
Laufer for their support. Our work is funded by the German Federal Ministry for Economic Affairs and Energy (BMWi) under grant agreement 01MD19003B (PLASS) and
01MK2008MD (Servicemeister).
\section{Bibliographical References}\label{reference}
\label{main:ref}

\bibliographystyle{lrec2022-bib}
\bibliography{citations}

\begin{thebibliography}{}

\bibitem[\protect\citename{Johnson \bgroup et al.\egroup }2016]{mimiciii}
Johnson, Alistair E.W. and Pollard, Tom J. and Shen, Lu and Lehman, Li-wei H.
  and Feng, Mengling and Ghassemi, Mohammad and Moody, Benjamin and Szolovits,
  Peter and Anthony Celi, Leo and Mark, Roger G.
\newblock (2016).
\newblock {\em {MIMIC}-{III}, a freely accessible critical care database}.

\bibitem[\protect\citename{Miranda{-}Escalada \bgroup et al.\egroup
  }2020]{codiesp}
Miranda{-}Escalada, A., Gonzalez{-}Agirre, A., Armengol{-}Estap{\'{e}}, J., and
  Krallinger, M.
\newblock (2020).
\newblock Overview of automatic clinical coding: Annotations, guidelines, and
  solutions for non-english clinical cases at codiesp track of {CLEF} ehealth
  2020.
\newblock In Linda Cappellato, et~al., editors, {\em Working Notes of {CLEF}
  2020 - Conference and Labs off the Evaluation Forum, Thessaloniki, Greece,
  September 22-25, 2020}, volume 2696 of {\em {CEUR} Workshop Proceedings}.
  CEUR-WS.org.

\end{thebibliography}


\begin{thebibliography}{}

\bibitem[\protect\citename{Artetxe and Schwenk}2019]{laser}
Artetxe, M. and Schwenk, H.
\newblock (2019).
\newblock Massively multilingual sentence embeddings for zero-shot
  cross-lingual transfer and beyond.
\newblock {\em Trans. Assoc. Comput. Linguistics}, 7:597--610.

\bibitem[\protect\citename{Artetxe \bgroup et al.\egroup
  }2020]{translationArtifacts}
Artetxe, M., Labaka, G., and Agirre, E.
\newblock (2020).
\newblock Translation artifacts in cross-lingual transfer learning.
\newblock In Bonnie Webber, et~al., editors, {\em Proceedings of the 2020
  Conference on Empirical Methods in Natural Language Processing, {EMNLP} 2020,
  Online, November 16-20, 2020}, pages 7674--7684. Association for
  Computational Linguistics.

\bibitem[\protect\citename{Asiimwe \bgroup et al.\egroup }2021]{data_silos}
Asiimwe, R., Lam, S., Leung, S., Wang, S., Wan, R., Tinker, A., McAlpine,
  J.~N., Woo, M. M.~M., Huntsman, D.~G., and Talhouk, A.
\newblock (2021).
\newblock From biobank and data silos into a data commons: convergence to
  support translational medicine.
\newblock {\em Journal of Translational Medicine}, 19(1):493, Dec.

\bibitem[\protect\citename{Carrino \bgroup et al.\egroup
  }2021]{spanish_clinicalbert}
Carrino, C.~P., Armengol{-}Estap{\'{e}}, J., Guti{\'{e}}rrez{-}Fandi{\~{n}}o,
  A., Llop{-}Palao, J., P{\`{a}}mies, M., Gonzalez{-}Agirre, A., and Villegas,
  M.
\newblock (2021).
\newblock Biomedical and clinical language models for spanish: On the benefits
  of domain-specific pretraining in a mid-resource scenario.
\newblock {\em CoRR}, abs/2109.03570.

\bibitem[\protect\citename{Cañete \bgroup et al.\egroup }2020]{SpanishBert}
Cañete, J., Chaperon, G., Fuentes, R., Ho, J.-H., Kang, H., and Pérez, J.
\newblock (2020).
\newblock Spanish pre-trained bert model and evaluation data.
\newblock In {\em PML4DC at ICLR 2020}.

\bibitem[\protect\citename{Chung \bgroup et al.\egroup
  }2020]{chung-etal-2020-improving}
Chung, H.~W., Garrette, D., Tan, K.~C., and Riesa, J.
\newblock (2020).
\newblock Improving multilingual models with language-clustered vocabularies.
\newblock In {\em Proceedings of the 2020 Conference on Empirical Methods in
  Natural Language Processing (EMNLP)}, pages 4536--4546, Online, November.
  Association for Computational Linguistics.

\bibitem[\protect\citename{Conneau and Lample}2019]{XLM}
Conneau, A. and Lample, G.
\newblock (2019).
\newblock Cross-lingual language model pretraining.
\newblock In Hanna~M. Wallach, et~al., editors, {\em Advances in Neural
  Information Processing Systems 32: Annual Conference on Neural Information
  Processing Systems 2019, NeurIPS 2019, December 8-14, 2019, Vancouver, BC,
  Canada}, pages 7057--7067.

\bibitem[\protect\citename{Conneau \bgroup et al.\egroup }2020]{XLMR}
Conneau, A., Khandelwal, K., Goyal, N., Chaudhary, V., Wenzek, G.,
  Guzm{\'{a}}n, F., Grave, E., Ott, M., Zettlemoyer, L., and Stoyanov, V.
\newblock (2020).
\newblock Unsupervised cross-lingual representation learning at scale.
\newblock In Dan Jurafsky, et~al., editors, {\em Proceedings of the 58th Annual
  Meeting of the Association for Computational Linguistics, {ACL} 2020, Online,
  July 5-10, 2020}, pages 8440--8451. Association for Computational
  Linguistics.

\bibitem[\protect\citename{Gu \bgroup et al.\egroup }2020]{pubmedbert}
Gu, Y., Tinn, R., Cheng, H., Lucas, M., Usuyama, N., Liu, X., Naumann, T., Gao,
  J., and Poon, H.
\newblock (2020).
\newblock Domain-specific language model pretraining for biomedical natural
  language processing.
\newblock {\em CoRR}, abs/2007.15779.

\bibitem[\protect\citename{Hu \bgroup et al.\egroup }2020]{Xtreme_benchmark}
Hu, J., Ruder, S., Siddhant, A., Neubig, G., Firat, O., and Johnson, M.
\newblock (2020).
\newblock {XTREME:} {A} massively multilingual multi-task benchmark for
  evaluating cross-lingual generalization.
\newblock {\em CoRR}, abs/2003.11080.

\bibitem[\protect\citename{Isbister \bgroup et al.\egroup
  }2021]{use_machine_translation}
Isbister, T., Carlsson, F., and Sahlgren, M.
\newblock (2021).
\newblock Should we stop training more monolingual models, and simply use
  machine translation instead?
\newblock In Simon Dobnik et~al., editors, {\em Proceedings of the 23rd Nordic
  Conference on Computational Linguistics, NoDaLiDa 2021, Reykjavik, Iceland
  (Online), May 31 - June 2, 2021}, pages 385--390. Link{\"{o}}ping University
  Electronic Press, Sweden.

\bibitem[\protect\citename{Koutsikakis \bgroup et al.\egroup }2020]{GreekBert}
Koutsikakis, J., Chalkidis, I., Malakasiotis, P., and Androutsopoulos, I.
\newblock (2020).
\newblock {GREEK-BERT:} the greeks visiting sesame street.
\newblock {\em CoRR}, abs/2008.12014.

\bibitem[\protect\citename{Labont{\'e} \bgroup et al.\egroup
  }2011]{Labont2011TheGI}
Labont{\'e}, R., Mohindra, K.~S., and Schrecker, T.
\newblock (2011).
\newblock The growing impact of globalization for health and public health
  practice.
\newblock {\em Annual review of public health}, 32:263--83.

\bibitem[\protect\citename{Le \bgroup et al.\egroup }2021]{Le2021LightweightAT}
Le, H., Pino, J.~M., Wang, C., Gu, J., Schwab, D., and Besacier, L.
\newblock (2021).
\newblock Lightweight adapter tuning for multilingual speech translation.
\newblock In {\em ACL/IJCNLP}.

\bibitem[\protect\citename{Li \bgroup et al.\egroup }2020]{Li2020BEHRTTF}
Li, Y., Rao, S., Solares, J. R.~A., Hassaine, A., Ramakrishnan, R., Canoy, D.,
  Zhu, Y., Rahimi, K., and Salimi-Khorshidi, G.
\newblock (2020).
\newblock Behrt: Transformer for electronic health records.
\newblock {\em Scientific Reports}, 10.

\bibitem[\protect\citename{Loshchilov and Hutter}2019]{AdamW}
Loshchilov, I. and Hutter, F.
\newblock (2019).
\newblock Decoupled weight decay regularization.
\newblock In {\em 7th International Conference on Learning Representations,
  {ICLR} 2019, New Orleans, LA, USA, May 6-9, 2019}. OpenReview.net.

\bibitem[\protect\citename{Mann and Yarowsky}2001]{yarowski}
Mann, G.~S. and Yarowsky, D.
\newblock (2001).
\newblock Multipath translation lexicon induction via bridge languages.
\newblock In {\em Second Meeting of the North {A}merican Chapter of the
  Association for Computational Linguistics}.

\bibitem[\protect\citename{Mulyar \bgroup et al.\egroup
  }2019]{pheno_bert_mulyar}
Mulyar, A., Schumacher, E., Rouhizadeh, M., and Dredze, M.
\newblock (2019).
\newblock Phenotyping of clinical notes with improved document classification
  models using contextualized neural language models.
\newblock {\em CoRR}, abs/1910.13664.

\bibitem[\protect\citename{Pfeiffer \bgroup et al.\egroup
  }2020a]{pfeiffer2020AdapterHub}
Pfeiffer, J., R{\"u}ckl{\'e}, A., Poth, C., Kamath, A., Vuli{\'c}, I., Ruder,
  S., Cho, K., and Gurevych, I.
\newblock (2020a).
\newblock Adapterhub: A framework for adapting transformers.
\newblock In {\em Proceedings of the 2020 Conference on Empirical Methods in
  Natural Language Processing: System Demonstrations}, pages 46--54.

\bibitem[\protect\citename{Pfeiffer \bgroup et al.\egroup }2020b]{madX}
Pfeiffer, J., Vulic, I., Gurevych, I., and Ruder, S.
\newblock (2020b).
\newblock {MAD-X:} an adapter-based framework for multi-task cross-lingual
  transfer.
\newblock In Bonnie Webber, et~al., editors, {\em Proceedings of the 2020
  Conference on Empirical Methods in Natural Language Processing, {EMNLP} 2020,
  Online, November 16-20, 2020}, pages 7654--7673. Association for
  Computational Linguistics.

\bibitem[\protect\citename{Ponti \bgroup et al.\egroup }2020]{Ponti2020XCOPAAM}
Ponti, E., Glavavs, G., Majewska, O., Liu, Q., Vuli'c, I., and Korhonen, A.
\newblock (2020).
\newblock Xcopa: A multilingual dataset for causal commonsense reasoning.
\newblock In {\em EMNLP}.

\bibitem[\protect\citename{Sch{\"{a}}fer and
  Friedrich}2020]{schafer_multilingual_icd}
Sch{\"{a}}fer, H. and Friedrich, C.~M.
\newblock (2020).
\newblock Multilingual {ICD-10} code assignment with transformer architectures
  using {MIMIC-III} discharge summaries.
\newblock In Linda Cappellato, et~al., editors, {\em Working Notes of {CLEF}
  2020 - Conference and Labs of the Evaluation Forum, Thessaloniki, Greece,
  September 22-25, 2020}, volume 2696 of {\em {CEUR} Workshop Proceedings}.
  CEUR-WS.org.

\bibitem[\protect\citename{Schneider \bgroup et al.\egroup
  }2020]{brasilian_biobert}
Schneider, E. T.~R., de~Souza, J. V.~A., Knafou, J., e~Oliveira, L. E.~S.,
  Copara, J., Gumiel, Y.~B., de~Oliveira, L. F.~A., Paraiso, E.~C., Teodoro,
  D., and Barra, C. M. C.~M.
\newblock (2020).
\newblock Biobertpt - {A} portuguese neural language model for clinical named
  entity recognition.
\newblock In Anna Rumshisky, et~al., editors, {\em Proceedings of the 3rd
  Clinical Natural Language Processing Workshop, ClinicalNLP@EMNLP 2020,
  Online, November 19, 2020}, pages 65--72. Association for Computational
  Linguistics.

\bibitem[\protect\citename{Sechidis \bgroup et al.\egroup
  }2011]{multilabel_stratified}
Sechidis, K., Tsoumakas, G., and Vlahavas, I.~P.
\newblock (2011).
\newblock On the stratification of multi-label data.
\newblock In Dimitrios Gunopulos, et~al., editors, {\em Machine Learning and
  Knowledge Discovery in Databases - European Conference, {ECML} {PKDD} 2011,
  Athens, Greece, September 5-9, 2011, Proceedings, Part {III}}, volume 6913 of
  {\em Lecture Notes in Computer Science}, pages 145--158. Springer.

\bibitem[\protect\citename{Soares and
  Krallinger}2019]{biomedical_translator_codie}
Soares, F. and Krallinger, M.
\newblock (2019).
\newblock {BSC} participation in the {WMT} translation of biomedical abstracts.
\newblock In Ondrej Bojar, et~al., editors, {\em Proceedings of the Fourth
  Conference on Machine Translation, {WMT} 2019, Florence, Italy, August 1-2,
  2019 - Volume 3: Shared Task Papers, Day 2}, pages 175--178. Association for
  Computational Linguistics.

\bibitem[\protect\citename{Solt \bgroup et al.\egroup }2009]{pheno_rules_solt}
Solt, I., Tikk, D., G{\'{a}}l, V., and Kardkov{\'{a}}cs, Z.~T.
\newblock (2009).
\newblock Research paper: Semantic classification of diseases in discharge
  summaries using a context-aware rule-based classifier.
\newblock {\em J. Am. Medical Informatics Assoc.}, 16(4):580--584.

\bibitem[\protect\citename{Tiedemann and Thottingal}2020]{opus_mt}
Tiedemann, J. and Thottingal, S.
\newblock (2020).
\newblock {OPUS-MT} - building open translation services for the world.
\newblock In Mikel~L. Forcada, et~al., editors, {\em Proceedings of the 22nd
  Annual Conference of the European Association for Machine Translation, {EAMT}
  2020, Lisboa, Portugal, November 3-5, 2020}, pages 479--480. European
  Association for Machine Translation.

\bibitem[\protect\citename{van Aken \bgroup et al.\egroup
  }2021]{clinical_outcome}
van Aken, B., Papaioannou, J., Mayrdorfer, M., Budde, K., Gers, F.~A., and
  L{\"{o}}ser, A.
\newblock (2021).
\newblock Clinical outcome prediction from admission notes using
  self-supervised knowledge integration.
\newblock In Paola Merlo, et~al., editors, {\em Proceedings of the 16th
  Conference of the European Chapter of the Association for Computational
  Linguistics: Main Volume, {EACL} 2021, Online, April 19 - 23, 2021}, pages
  881--893. Association for Computational Linguistics.

\bibitem[\protect\citename{Wang \bgroup et al.\egroup }2019]{pheno_cnn_wang}
Wang, Y., Sohn, S., Liu, S., Shen, F., Wang, L., Atkinson, E.~J., Amin, S., and
  Liu, H.
\newblock (2019).
\newblock A clinical text classification paradigm using weak supervision and
  deep representation.
\newblock {\em {BMC} Medical Informatics Decis. Mak.}, 19(1):1:1--1:13.

\bibitem[\protect\citename{Wolf \bgroup et al.\egroup }2020]{huggingface}
Wolf, T., Debut, L., Sanh, V., Chaumond, J., Delangue, C., Moi, A., Cistac, P.,
  Rault, T., Louf, R., Funtowicz, M., Davison, J., Shleifer, S., von Platen,
  P., Ma, C., Jernite, Y., Plu, J., Xu, C., Le~Scao, T., Gugger, S., Drame, M.,
  Lhoest, Q., and Rush, A.
\newblock (2020).
\newblock Transformers: State-of-the-art natural language processing.
\newblock In {\em Proceedings of the 2020 Conference on Empirical Methods in
  Natural Language Processing: System Demonstrations}, pages 38--45, Online,
  October. Association for Computational Linguistics.

\bibitem[\protect\citename{Y-Hassan and Tornvall}2018]{takotsubo}
Y-Hassan, S. and Tornvall, P.
\newblock (2018).
\newblock Epidemiology, pathogenesis, and management of takotsubo syndrome.
\newblock {\em Clinical autonomic research : official journal of the Clinical
  Autonomic Research Society}, 28(1):53--65, Feb.
\newblock PMC5805795[pmcid].

\bibitem[\protect\citename{Yao \bgroup et al.\egroup }2019]{pheno_cnn_yao}
Yao, L., Mao, C., and Luo, Y.
\newblock (2019).
\newblock Clinical text classification with rule-based features and
  knowledge-guided convolutional neural networks.
\newblock {\em {BMC} Medical Informatics Decis. Mak.}, 19-S(3):31--39.

\bibitem[\protect\citename{Zhang \bgroup et al.\egroup
  }2021]{zhang-etal-2021-self}
Zhang, J., Bolanos~Trujillo, L., Li, T., Tanwar, A., Freire, G., Yang, X., Ive,
  J., Gupta, V., and Guo, Y.
\newblock (2021).
\newblock Self-supervised detection of contextual synonyms in a multi-class
  setting: Phenotype annotation use case.
\newblock In {\em Proceedings of the 2021 Conference on Empirical Methods in
  Natural Language Processing}, pages 8754--8769, Online and Punta Cana,
  Dominican Republic, November. Association for Computational Linguistics.

\end{thebibliography}

\section{Language Resource References}
\label{lr:ref}
\bibliographystylelanguageresource{lrec2022-bib}
\bibliographylanguageresource{languageresource}

\end{document}